\documentclass[sigconf]{acmart}

\AtBeginDocument{%
  \providecommand\BibTeX{{%
    \normalfont B\kern-0.5em{\scshape i\kern-0.25em b}\kern-0.8em\TeX}}}

\setcopyright{rightsretained}
\acmConference[SIGIR eCom'20]{ACM SIGIR Workshop on eCommerce}{July 30, 2020}{Virtual Event, China}
\acmYear{2020}
\copyrightyear{2020}
\makeatletter
\renewcommand\@formatdoi[1]{\ignorespaces}
\makeatother
\acmISBN{}
\usepackage{array,booktabs,ragged2e}
\newcolumntype{R}[1]{>{\RaggedLeft\arraybackslash}p{#1}}

\begin{document}

\title{Discriminative Pre-training for Low Resource Title Compression in Conversational Grocery}


\author{Snehasish Mukherjee}
\affiliation{%
  \institution{Walmart Labs}
  \streetaddress{8600 Datapoint Drive}
  \city{Sunnyvale}
  \state{California}
  \postcode{78229}}
\email{smukherjee@walmartlabs.com}

\author{Phaniram Sayapaneni}
\affiliation{%
  \institution{Walmart Labs}
  \streetaddress{8600 Datapoint Drive}
  \city{Sunnyvale}
  \state{California}
  \postcode{78229}}
\email{phaniram.sayapaneni@walmartlabs.com}

\author{Shankar Subramanya}
\affiliation{%
  \institution{Walmart Labs}
  \streetaddress{8600 Datapoint Drive}
  \city{Sunnyvale}
  \state{California}
  \postcode{78229}}
\email{ssubramanya@walmartlabs.com}



\renewcommand{\shortauthors}{Mukherjee, Sayapaneni and Subramanya}

\begin{abstract}
  The ubiquity of smart voice assistants has made conversational shopping commonplace. This is especially true for low consideration segments like grocery. A central problem in conversational grocery is the automatic generation of short product titles that can be read out fast during a conversation. Several supervised models have been proposed in the literature that leverage manually labeled datasets and additional product features to generate short titles automatically. However, obtaining large amounts of labeled data is expensive and most grocery item pages are not as feature-rich as other categories. To address this problem we propose a pre-training based solution that makes use of unlabeled data to learn contextual product representations which can then be fine-tuned to obtain better title compression even in a low resource setting. We use a self-attentive BiLSTM encoder network with a time distributed softmax layer for the title compression task. We overcome the vocabulary mismatch problem by using a hybrid embedding layer that combines pre-trained word embeddings with trainable character level convolutions. We pre-train this network as a discriminator on a replaced-token detection task over a large number of unlabeled grocery product titles. Finally, we fine tune this network, without any modifications, with a small labeled dataset for the title compression task. Experiments on Walmart’s online grocery catalog show our model achieves performance comparable to state-of-the-art models like BERT and XLNet. When fine tuned on all of the available training data our model attains an F1 score of 0.8558 which lags the best performing model, BERT-Base, by 2.78\% and XLNet by 0.28\% only, while using 55 times lesser parameters than both. Further, when allowed to fine tune on 5\% of the training data only, our model outperforms BERT-Base by 24.3\% in F1 score.
\end{abstract}




\maketitle

\section{Introduction}

Voice commerce is a rapidly growing vertical, with an estimated market size of \$2 Billion already and forecasts for hitting \$40 Billion+ by 2022 \cite{DBLP:conf/sigir/XiaoM19}. Moreover, voice is being touted as the dominant channel for driving customer engagement in the next decade. As such, conversational commerce, a broader term often used interchangeably with voice commerce, is an area of strategic priority for most e-commerce businesses. From a user adoption and repeat usage standpoint, the prospects for voice commerce are even more exciting for low consideration segments like Grocery.

\begin{figure}[h]
  \centering
  \includegraphics[width=\linewidth]{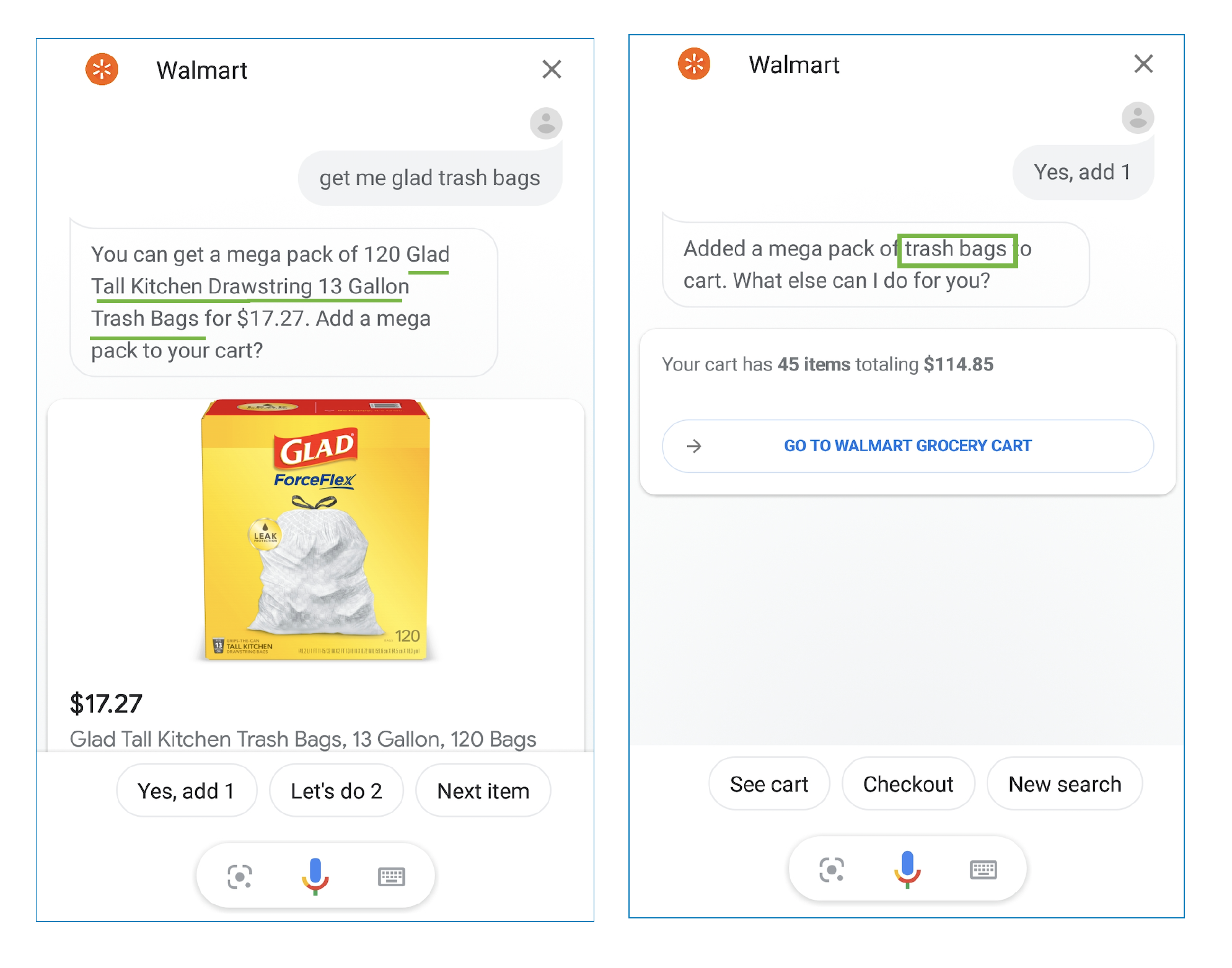}
  \caption{Walmart Voice Order bot for online grocery. Though search response is fairly detailed, the add to cart response has the maximum possible compression of the product title.}
  \Description{A woman and a girl in white dresses sit in an open car.}
  \label{wmt:voiceorder}
\end{figure}

This big potential notwithstanding, the user experience for voice shopping, and conversational grocery in particular, is still nascent. One such suboptimality is the reading out of long product titles during a conversation between an artificial shopping agent and a customer. As noted in \cite{DBLP:conf/cikm/SunJSPOW18}, product titles in e-commerce catalogs are made very long and informative with Search Engine Optimization in mind. However, this translates to a poor grocery voice shopping experience where transactions are mostly repeat purchases and users are well aware of the products that are being added to cart. Reading out a product title like ``Glad OdorShield Tall Kitchen Drawstring Trash bags - Febreze Fresh Clean - 13 Gallon - 40 count (Packaging May Vary)'' for 7 to 10 seconds during an add-to-cart response, for example, is an unnecessary test of user's patience and can lead to greater drop out rates before checkout. Instead, a succinct summarization of the title, like ``trash bags'', is more appropriate in this scenario. From a voice UX standpoint, any product title with $>= 5$ words is considered to be long, while those with $<= 4$ words are considered short and voice friendly. This necessitates exploring automatic product title summarization techniques that can take in a long product title and convert it into a short and voice friendly title.

 Automatic generation of short product titles have been studied in the context of voice and mobile shopping in \cite{DBLP:conf/aaai/WangTQLLSL18, DBLP:conf/cikm/SunJSPOW18, DBLP:journals/corr/abs-1811-04498, DBLP:conf/sigir/XiaoM19, DBLP:conf/aaai/GongLZOLD19}. All proposed methods are supervised learning models that leverage labeled training data and other product features to compress titles. Much better results can be obtained, if this labeled data is used to fine tune state-of-the-art pre-trained models like BERT,  XLNet, ELECTRA \cite{DBLP:conf/iclr/ClarkLLM20} etc. However, this poses a few problems in the context of conversational grocery. Firstly, labeled data is expensive to obtain and grocery product pages and catalog are not as feature-rich as some other segments like fashion and furniture, for example. Secondly, heavily parameterized models like BERT and XLNet that require more powerful GPUs to train, do not make a strong case in favor of cost-effectiveness, which is a carefully tracked metric for low margin segments like grocery.

We try to address these challenges in this paper by proposing a lightweight self-attentive BiLSTM architecture that can be pre-trained on unlabeled long product titles to obtain performance comparable to state-of-the-art models for the title summarization task. Since we are low on training data, we adopt the discriminative pre-training strategy, recently introduced in \cite{DBLP:conf/iclr/ClarkLLM20}, for our model. Pre-training models as discriminators on a replaced token detection task results in higher sample efficiency and has been shown to outperform masked language model based pre-training. Experiments on Walmart's online grocery catalog show that our pre-trained model achieves performance comparable to BERT, XLNet, and ELECTRA while using 55X and 7X lesser parameters respectively. Further, when allowed to fine tune on less than 20\% of the available training data, our pre-trained model outperforms all three.

\section{Related Work}
\subsection{Text Summarization}
Our problem can be classified under the broad category of text summarization. This area has evolved quite a bit with the advent of RNNs.  \cite
{DBLP:books/sp/mining2012/NenkovaM12, DBLP:conf/icetet/DalalM13, DBLP:journals/corr/AllahyariPASTGK17, DBLP:journals/air/GambhirG17} present comprehensive  surveys of neural as well as classical text summarization techniques. While document level summarization \cite{DBLP:conf/acl/0001L16, DBLP:conf/acl/BingLLLGP15, DBLP:conf/aaai/NallapatiZZ17} deals with the problem of generating document level summaries of the content, our work is best described as sentence compression \cite{DBLP:conf/acl/WangRCFC13, DBLP:conf/pacling/LaiSM17, DBLP:conf/naacl/ChopraAR16, DBLP:conf/emnlp/MiaoB16, DBLP:conf/acl/ZhaoLA18, DBLP:conf/conll/FevryP18, DBLP:conf/emnlp/FilippovaACKV15} which involves summarizing long sentences to shorter ones while preserving the core intent. 

There are 2 distinct flavors of summarization independent of the source granularity; \textit{Abstractive} and \textit{Extractive}. Abstractive summarization \cite{DBLP:conf/naacl/ChopraAR16, DBLP:conf/conll/NallapatiZSGX16,DBLP:journals/corr/abs-1812-02303, DBLP:conf/acl/BingLLLGP15} can produce summaries consisting of words not present in the input, while Extractive summarization \cite{DBLP:conf/emnlp/FilippovaACKV15,DBLP:conf/conll/FevryP18, DBLP:conf/acl/ZhaoLA18, DBLP:conf/pacling/LaiSM17} aims to generate summaries using words or sentences extracted from the original input. We will focus on neural approaches for extractive flavor of sentence compression for the remainder of this section.

A popular approach to extractive sentence compression is to model the problem as a sequence-to-sequence learning problem using an encoder-decoder based architecture \cite{DBLP:conf/emnlp/FilippovaACKV15, DBLP:conf/conll/FevryP18, DBLP:conf/emnlp/MiaoB16}. The encoder learns a distributed representation of the input sentence which is then fed to the decoder which is trained to produce a binary 0/1 label for each input word denoting whether to delete or keep that word in the output summary. Beam search can be used \cite{DBLP:conf/emnlp/FilippovaACKV15, DBLP:conf/acl/WangRCFC13, DBLP:conf/emnlp/WisemanR16} to sample from the decoder output probability distribution to generate the most likely compression. 

\cite{DBLP:conf/pacling/LaiSM17} and \cite{DBLP:conf/acl/ZhaoLA18} present 2 approaches different from the encoder-decoder based approaches. \cite{DBLP:conf/acl/ZhaoLA18} uses Reinforcement Learning with KEEP and DELETE policies. Words are kept or deleted based on the policy while a pre trained language model provides feedback on the generated summary. This results in the system learning the optimal policy over time. Like all the other deletion based compression models, \cite{DBLP:conf/pacling/LaiSM17} still tags words in the input sequence with a 0/1 label, but treats it as a sequence labeling problem, instead of a sequence generation problem. Hence it does away with the decoder and employs stacked BiLSTM layers with a final CRF layer at the output that does the 0/1 classification. Our current work closely follows this approach with minor modifications.

Most of the techniques discussed here, with the exception of \cite{DBLP:conf/pacling/LaiSM17}, require a considerable amount of training data. While \cite{DBLP:conf/emnlp/MiaoB16, DBLP:conf/conll/FevryP18, DBLP:conf/acl/ZhaoLA18} are trained on the Gigaword corpus, \cite{DBLP:conf/emnlp/FilippovaACKV15} is trained on 2 Million news headline and summary pairs. Though these 2 Million examples were synthetically generated following the method in \cite{DBLP:conf/emnlp/FilippovaA13}, it is based on the syntactic structure of the English language and its parse trees. \cite{DBLP:conf/conll/FevryP18} also proposes an unsupervised approach involving generation of training data by adding noise, but the results lag behind supervised approaches. Since the distribution of the words in the product titles, and in general the vocabulary of our problem domain, is different from general English, data sparsity and unavailability of e-commerce specific embeddings pose a challenge for us. Interestingly, \cite{DBLP:conf/pacling/LaiSM17} reported good performance with only 10000 pairs of original and compressed sentences, which is the reason we chose to implement a modified version of this approach to solve our problem of product title compression.
\subsection{Title Summarization in E-commerce}
Product title summarization in the context of voice and mobile shopping has been  studied in \cite{DBLP:conf/aaai/WangTQLLSL18, DBLP:conf/cikm/SunJSPOW18,DBLP:journals/corr/abs-1811-04498, DBLP:conf/sigir/XiaoM19, DBLP:conf/aaai/GongLZOLD19}. Amongst them, \cite{DBLP:conf/aaai/GongLZOLD19} defined title summarization as a sequence classification problem, where a binary decision is made at each word, given a long product title. A feature vector has been applied, comprising term frequency (tf) and inverse document frequency (idf) for each word. However, in this approach, large training data (500,000 samples) has been used and the results didn't prove to be significantly better than a simple BiLSTM approach. Another interesting approach has been proposed, where the problem has been defined as multi-task learning objective \cite{DBLP:conf/aaai/WangTQLLSL18} to compress the product title using user search log data. The multi-task learning objective involves two networks, one network to select the most informative words from the product title as a compressed title and the other network to generate the user search query from the product title. These two networks have been constrained to share the same product title encoder and additionally, the attention distributions from these two networks were constrained to agree with each other. However, in this approach, the size of the training data is large (185,386 samples) and additionally requires user search data. Another recent approach has been proposed in which the product title summarization has been framed as a Binary Named Entity Recognition problem \cite{DBLP:conf/sigir/XiaoM19}. The model architecture involves a simple bi-directional LSTM encoder/decoder
network (2 layer LSTMs) with an attention mechanism. ANOVA and post-hoc tests showed that with this approach, there was no statistical difference between model outputs and human-labeled short titles.
 
\section{DATASET}

Our main dataset consists of product titles, and their corresponding human generated summaries for \textbf{40,445} top selling Walmart grocery products during the calendar year 2018. Crowd workers were given the product titles and asked to generate their short summaries by choosing words to retain from the original title. Product description, brand name and category information were also provided to help workers decide on the most salient tokens for unfamiliar items. The task was intentionally vague about the target length of the summary so as to not introduce any bias. Examples provided were minimal identity preserving compression, with extra terms added rarely to improve fluency. Table \ref{tab:sample} lists some samples from this dataset.   We call this dataset the \textbf{title summary dataset}. Unlike \cite{DBLP:conf/sigir/XiaoM19} our short titles do not necessarily contain additional entities like brand, size, pack descriptors etc that can be easily obtained from structured catalog data.

\begin{table}
  \caption{Crowd Generated Short Titles}
  \label{tab:sample}
  \begin{tabular}{p{0.03\textwidth}p{0.30\textwidth}p{0.10\textwidth}}
    \toprule
    No.&Long Title&Short Title\\
    \midrule
    1.&Freshness Guaranteed Sliced Fruit Cake, 13 oz & Fruit Cake\\
    2.&OGX Hydrating + Teatree Mint Conditioner Salon, 25.4oz & Conditioner\\
    3.&Child of Mine by Carter's Places and Spaces 3 Pocket Duffle Diaper Bag Gray & Diaper Bag\\
    4.&Mainstays Stall Size 54" x 78" Medium Weight PEVA Shower Liner, 1 Each&Shower Liner\\
  \bottomrule
\end{tabular}
\end{table}

\begin{table}
  \caption{Dataset Statistics}
  \label{tab:stats}
  \begin{tabular}{p{0.30\textwidth}p{0.12\textwidth}}
    \toprule
    Metric&Value\\
    \midrule
    No. of pre-training samples & 256,298\\
    Shortest/Median/Longest sample lengths & 3/10/35 words\\
    Word Vocabulary size & 67,634\\
    Character Vocabulary size & 69\\
    Words missing from embedding (UNK) & 17041 (25.2\%)\\
    \midrule
    No. of title compression samples & 40,445\\
    Shortest/Median/Longest short titles & 1/2/5 tokens\\
    Shortest/Median/Longest long titles & 4/10/35 tokens\\
  \bottomrule
\end{tabular}
\end{table}

We use an additional unlabeled dataset of around \textbf{256,298} long product titles of items published in Walmart's online grocery catalog, which is used for pre training the network. We refer to this dataset as the \textbf{product titles dataset}. Table \ref{tab:stats} lists the statistics for both these datasets. The length distribution across the crowd generated summaries and the original product titles conforms well to the UX requirement which considers product titles with $>= 5$ words as long and those with $<= 4$ words as short and voice friendly.

\section{Model}

We model the problem as a binary sequence labelling problem, where each element of the input sequence is assigned the label \textbf{0} or \textbf{1}. The sequence elements labeled \textbf{1}, taken in order, forms the short title. For the sequence labelling problem we use a 3 layer architecture consisting of an embedding layer, an encoder layer, and a final point-wise classification layer.
\paragraph{Embedding layer}
 A major problem in applying pre-trained embeddings to a specific domain like retail is vocabulary mismatch where many private labels, brands, pack descriptors etc are treated as unknown words. To overcome this we use a combination of fixed pre-trained word embeddings and randomly initialized, trainable, character level embeddings as described in \cite{DBLP:conf/iclr/SeoKFH17}. Hence, our model takes 2 inputs; $\mathbf{x}_{w} \in \mathbb{Z}^{N}$ which is a vector containing indices of words in the input product title, and $\mathbf{x}_{c} \in \mathbb{Z}^{N \times C}$ which contains the indices of the characters in each word, where N is the maximum sequence length and C is the maximum word length. We use character level convolutions (CharCNN) \cite{kim-2014-convolutional} on $\mathbf{x}_{c}$ to combine and project the character level embeddings for each word onto $\mathbb{R}^{e_{char}}$. We combine these two word embeddings using a highway network \cite{DBLP:journals/corr/SrivastavaGS15} to obtain the final embedding $\mathbf{x}_{emb}$. 
\begin{eqnarray}
    \mathbf{x}_{wemb}^{i}&=&\operatorname{word-embedding}\hspace{3pt}(\mathbf{x}_{w}^{i}), \hspace{3pt}  \in \mathbb{R}^{e_{word}}\\
    \mathbf{x}_{cemb}^{i}&=&\operatorname{CharCNN}\hspace{3pt}(\mathbf{x}_{c}^{i}), \hspace{3pt}  \in \mathbb{R}^{e_{char}}\\
    \mathbf{x}_{emb}^{i}&=&\operatorname{highway}\hspace{3pt}([\mathbf{x}_{cemb}^{i};\mathbf{x}_{wemb}^{i}]), \hspace{3pt} \in \mathbb{R}^{e_{char} + e_{word}}
\end{eqnarray}

Instead of repeating the details of how these components work we direct the reader to \cite{DBLP:conf/iclr/SeoKFH17, DBLP:journals/corr/SrivastavaGS15, kim-2014-convolutional} for further details.
\paragraph{Encoder layer} The encoder layer uses 3 stacked layers of bidirectional LSTMs to obtain contextualized representation $\mathbf{x}_{b}^{i} \in \mathbb{R}^{2 h}$ for the $i_{th}$ sequence element as the concatenation of the hidden states, each of dimension $h$, from the forward and backward passes of the LSTM units in the 3rd layer 

\begin{equation}
    \mathbf{x}_{b}^{i}=\left[\mathbf{h}_{f}^{(i)[3]};\mathbf{h}_{b}^{(i)[3]}\right], i \in \{1, 2, ... N\}
\end{equation}

We further augment this contextualized representation $\mathbf{x}_{b}^{i}$ of each sequence element by using multiplicative self attention to jointly attend to all other sequence elements $\mathbf{x}_{b}^{j}, j \in \{1, 2, ... N\}$ without having to go through any gating mechanism. We thus obtain the final encoding $\mathbf{x}_{enc}^{i}$ for each sequence element as follows

\begin{eqnarray}
    \mathbf{e}_{ij}&=&\mathbf{x}_b{}^i{}^{\top} \mathbf{W}_{s} \mathbf{x}_b{}^j, \hspace{5pt} \mathbf{e}_{ij} \in \mathbb{R}\\[5pt]
    \mathbb{\alpha}_{ij}&=&\frac{\exp \left(\mathbf{e}_{ij}\right)}{\sum_{k=1}^{N} \exp \left(\mathbf{e}_{i k}\right)}\\[5pt]
    \mathbf{x}_{enc}^{i}&=&\sum_{j=1}^{N} \alpha_{i j} \mathbf{x}_{b}^{i}
\end{eqnarray}

\noindent where $\mathbf{W}_{s} \in \mathbb{R}^{2 h \times 2 h}$ is a trainable parameter matrix and $\alpha_{ij}$ determines the contribution of the $j_{th}$ sequence element in computing the representation for the $i_{th}$ sequence element.

\paragraph{Classification layer} In the final layer we project the contextualized embeddings for each sequence element as obtained from the encoder layer to $\mathbb{R}^{2}$ using a point-wise fully connected layer, parameterized by the weight matrix $\mathbf{W_c} \in \mathbb{R}^{2 h \times 2}$ and the bias $\mathbf{b_c} \in  \mathbb{R}^{2}$, which when operated upon by a softmax operator yields $\mathbf{y}_i$, the probability distribution across the output class labels for the $i_{th}$ sequence element. Succinctly,
\begin{equation}
\mathbf{y}_{i}=\operatorname{softmax}\left(\mathbf{W}_{c}^{T}\mathbf{x}_{enc}^{i} + \mathbf{b_c}\right), i \in \{1, 2, ... N\}
\end{equation}

\paragraph{Training} We train our model to minimize the weighted binary cross entropy loss $L(\theta)$ given by 
\begin{equation}
  L(\theta)=-\frac{1}{N} \sum_{i}^{N} \alpha \cdot \hat{\mathbf{y}}_{i}\log \left(\mathbf{y}_{i}\right)+\beta \cdot (1 - \hat{\mathbf{y}}_{i})\log (1 - \mathbf{y}_{i})
\end{equation}

\noindent where $N$ is the sequence length, $\mathbf{y}_{i}$ is the probability that the $i_{th}$ sequence element belongs to class \textbf{1}, $\hat{\mathbf{y}}_{i}$ is the ground truth label, $\alpha$ is the weight for class \textbf{0}, and $\beta=1-\alpha$, is the weight for class \textbf{1}. We choose $\alpha=0.1$ and hence $\beta=0.9$ since roughly $\frac{9}{10}$ of all the token labels are \textbf{0}

Our model architecture is similar to \cite{DBLP:conf/emnlp/FilippovaACKV15, DBLP:conf/pacling/LaiSM17, DBLP:conf/sigir/XiaoM19} with some key differences. Firstly, unlike our hybrid embedding layer, none of the aforementioned solutions have any mechanism to address the vocabulary mismatch problem between pre-trained embeddings and the training corpus. Next, unlike \cite{DBLP:conf/pacling/LaiSM17, DBLP:conf/sigir/XiaoM19} that use encoder-decoder based architecture, ours in an encoder-only architecture. Further, \cite{DBLP:conf/emnlp/FilippovaACKV15} uses left to right LSTM only and hence requires feeding the input in reverse to condition on the right context, which we achieve using BiLSTM. Finally, unlike \cite{DBLP:conf/emnlp/FilippovaACKV15, DBLP:conf/pacling/LaiSM17, DBLP:conf/sigir/XiaoM19} we attend to the global context while encoding each sequence position using a self attention layer.

\section{Pre-training}

As in \cite{DBLP:conf/iclr/ClarkLLM20}, we pre-train our network as a discriminator on a replaced token detection task. We corrupt each long title in the product titles dataset, by randomly selecting some fraction $f$ of its tokens and replacing them with another token. Though experiments in \cite{DBLP:conf/naacl/DevlinCLT19} show that best results are obtained with 15\% of the tokens corrupted, we choose $f =0.25$ since our corpus is much smaller. To ensure that the network gets a chance to make predictions for all positions, we repeat the token replacement process multiple times for the same long title until the corruption procedure covers all tokens. This results in multiple copies of the same title, with tokens replaced in mutually disjoint positions. We generate a binary sequence label for each corrupted line that has \textbf{1} for the positions that were replaced and \textbf{0} everywhere else. In order to not bias the network towards predicting at least one \textbf{1} label in each input, we also include the original, uncorrupted copy of each product title. This procedure applied on our dataset results in a corpus of ~1.27 Million product titles for supervised pre-training of the network. We use the same weighted binary cross entropy loss function as in equation 9. Since the median number of tokens per title in our corpus is around 10, with $f=0.25$ and $N=35$ we used $\alpha=\frac{10f}{N}\approx0.07$ and $\beta=1-\alpha=0.93$. We train this network for 4 epochs which results in accuracy of 0.9897 on the replaced token detection task.

\subsection{Token Replacement} Intuitively, higher the quality of the replaced token, harder it is for the network to guess that it is replaced, and hence better is the latent representations learnt during the pre-training process. Unlike \cite{DBLP:conf/iclr/ClarkLLM20} which samples the replacement tokens from a small masked language model trained jointly, but not adversarially, with the network, we obtain our replacement tokens from a skip gram model that maximizes the log likelihood in a window centered on the token to be replaced. This allows us to improve our compute efficiency while obtaining reasonably good candidate replacement tokens. If $w_i$ is the token at the $i_{th}$ position that is to be replaced, then we choose $w_r$, the replacement token as
\begin{displaymath}
  w_r = \underset{w \in V^{'}}{\operatorname{argmin}} \sum_{k=-n}^{n}-\log P_{s}(w|w_{i+k}; l_w)
\end{displaymath}
where $V$ is the vocabulary, $V^{'}=V - \{w_{i+k}; -n \leq k \leq +n\}$, $l_w=2n+1$ is the window size, and $P_s(w_i|w_j; l_w)$ is the conditional distribution for occurrence of $w_i$ in a window of length $l_w$ centered on $w_j$.

\section{Experiments}
We implement our model and it's various ablations in the Tensorflow framework\footnote{https://www.tensorflow.org/}. For our baseline models, we use their Pytorch implementations available at HuggingFace\footnote{https://huggingface.co/transformers/}. We minimize the model loss as given in equation 9 using the Adam optimizer. We do not tune any hyper parameters and use the default settings for Adam throughout with $lr=0.001$, $\beta_1=0.9$, and $\beta_2=0.999$. We use the same dropout probability of 0.2 between all layers. All layer weights, with the exception of the highway layer and the character embeddings, use Xavier normal initialization and biases are set to 0. All our models are trained on a single Nvidia V100 GPU for 15 epochs, or 1 hour, or until convergence, whichever is earlier. We define convergence as 3 consecutive epochs without any improvements in metrics on the validation set. For fine-tuning our pre-trained model, we use gradual unfreezing of layers in a top down manner as in \cite{DBLP:conf/acl/RuderH18}, but we get the best results when we keep our learning rate fixed at 0.001 across all layers and batches. We evaluate our model performance on a held out test set consisting of 8089 human generated short titles. We report 2 different metrics for our models; ROUGUE-1 F1 score and Exact Match percentage, henceforth referred to as \textbf{F1} and \textbf{EM} respectively. EM, as the name suggests, refers to the percentage of outputs that exactly match the human generated title compression. For comparing model performances we use the \textbf{F1} score only. 

\subsection{Dataset Preparation}
We minimally normalize our datasets by converting everything to lower case and squeezing consecutive white space characters. Given the nature of the Walmart catalog data, 2 additional normalization steps were necessary; converting all ``\&'' symbols to ``and'' and padding commas with whitespaces so that they are treated as separate tokens. We follow a simple tokenization scheme where we split each product title by whitespace. We keep the maximum sequence length at 35, truncating longer product titles and padding shorter titles with the PAD token. The maximum token length we use is 15, with similar padding and truncating scheme. We extract our word and character vocabularies from the product titles dataset instead of the title summary dataset since the former is a super set of the later. Our corpus has a word vocabulary size of 67634 including the ``UNK'' and ``PAD'' tokens, while the character vocabulary size stands at 69. We set aside 20\% of the human generated title summary dataset as our test set, 8\% as validation set and the remaining 72\% as our training set. 

\subsection{Ablations}

We perform several ablations, as well as some additions, on the model proposed in section 4 to underscore the effect of each of the components. We name our model CB3SA, choosing to retain the first character for each layer (CharCNN, BiLSTM(3 stacks) and self attention). Additions and ablations are denoted with a + and - operator respectively after the model name. For example CB3SA+PT refers to the pre-trained version of our model, while CB3SA-CharCNN refers to the version of our model with the character level word embedding layer removed. Table \ref{tab:ablation} lists the various combinations we tried and the results on the test set.

\begin{table}
  \caption{Ablations}
  \label{tab:ablation}
  \begin{tabular}{p{0.20\textwidth}p{0.10\textwidth}p{0.10\textwidth}}
    \toprule
    \textbf{Model}&\textbf{F1}&\textbf{EM}\\
    \midrule
    CB3SA          & 0.8465 & 62.24\\
    \midrule
    CB3SA+PT       & \textbf{0.8558} & \textbf{63.83}\\
    \midrule
    CB3SA-CharCNN  & 0.8414 & 60.13\\
    \midrule
    CB3SA-BLSTM1   & 0.8455 & 62.37\\
    \midrule
    CB3SA-SA       & 0.8417  & 60.22\\
    \midrule
    CB3SA-SA+NWSA7 & 0.8458 & 62.39\\
    \midrule
    CB3SA-SA+MHSA8 & 0.8420 & 59.72\\
  \bottomrule
\end{tabular}
\end{table}

Clearly, the pre-training process (model CB3SA+PT) contributed a significant boost in performance over the non pretrained model (CB3SA) while removing the charCNN layer (CB3SA-CharCNN) causes the most significant drop in performance. Apart from these, there are a few interesting observations. Firstly, removing a BiLSTM layer (model CB3SA-BLSTM1) has one of the least negative impacts together with reduced training and inference times. Secondly, removing the vanilla self attention layer and adding a multi-headed self attention layer with 8 attention heads (model CB3SA-SA+MHSA8) proves to be as detrimental as removing the self attention layer altogether (model CB3SA-SA). Finally, using a narrow width attention with window length of 7 instead of our global self attention (model CB3SA-SA+NWSA7) causes the least drop in performance.

\subsection{Baselines}

We fine tune our pre-trained model (CB3SA+PT) on the entire training dataset and compare it's performance against several strong baselines; XLNet, BERT, RoBERTa,  DistillBERT, and ELECTRA - all fine tuned on the same training set. Table \ref{tab:baselines} lists the results. Additionally, the F1 column shows the percentage by which our F1 lags the respective models and the Params column shows the factor by which our parameter set is smaller.

\begin{table}
  \caption{Comparison against SOTA models}
  \label{tab:baselines}
  \begin{tabular}{p {0.10\textwidth}  p {0.10\textwidth}  p {0.14\textwidth} p {0.06\textwidth}}
    \toprule
    \textbf{Model}&\textbf{Params}&\textbf{F1}&\textbf{EM}\\
    \midrule
    CB3SA+PT      &   \textbf{2M} & \textbf{0.8558} & 63.83\\
    \midrule
    XLNet         & 110M (55X) & 0.8582 (-0.28\%) & 74.25\\
    \midrule
    BERT-Base     & \textbf{110M (55X)} & \textbf{0.8803 (-2.78\%)} & 69.17\\
    \midrule
    RoBERTa       & 125M (62X) & 0.7644 (+11.96\%)& 58.17\\
    \midrule
    ELECTRA & 14M (7X)   & 0.8689 (-1.50\%)& 66.48\\
    \midrule
    Distill BERT  & 66M (33X)  & 0.8707 (-1.71\%)& 67.18\\
  \bottomrule
\end{tabular}
\end{table}
\textbf{Our model is able to use 55X lesser parameters than BERT and yet obtain a comparable performance, lagging BERT Base by 2.78\% of F1 score on the test set.} Among the SOTA models, ELECTRA-Small seems to provide the best parameter efficiency by uisng 7.8X lesser parameters than BERT-Base while lagging by only 1.3\% on F1 score on the test set. Interestingly, RoBERTa, the biggest model that we tried with 125M parameters, under performed by a huge margin, lagging our F1 score by 11.96\%.

\subsection{Low Resource Setting}

\begin{figure}[h]
  \centering
  \includegraphics[width=\linewidth]{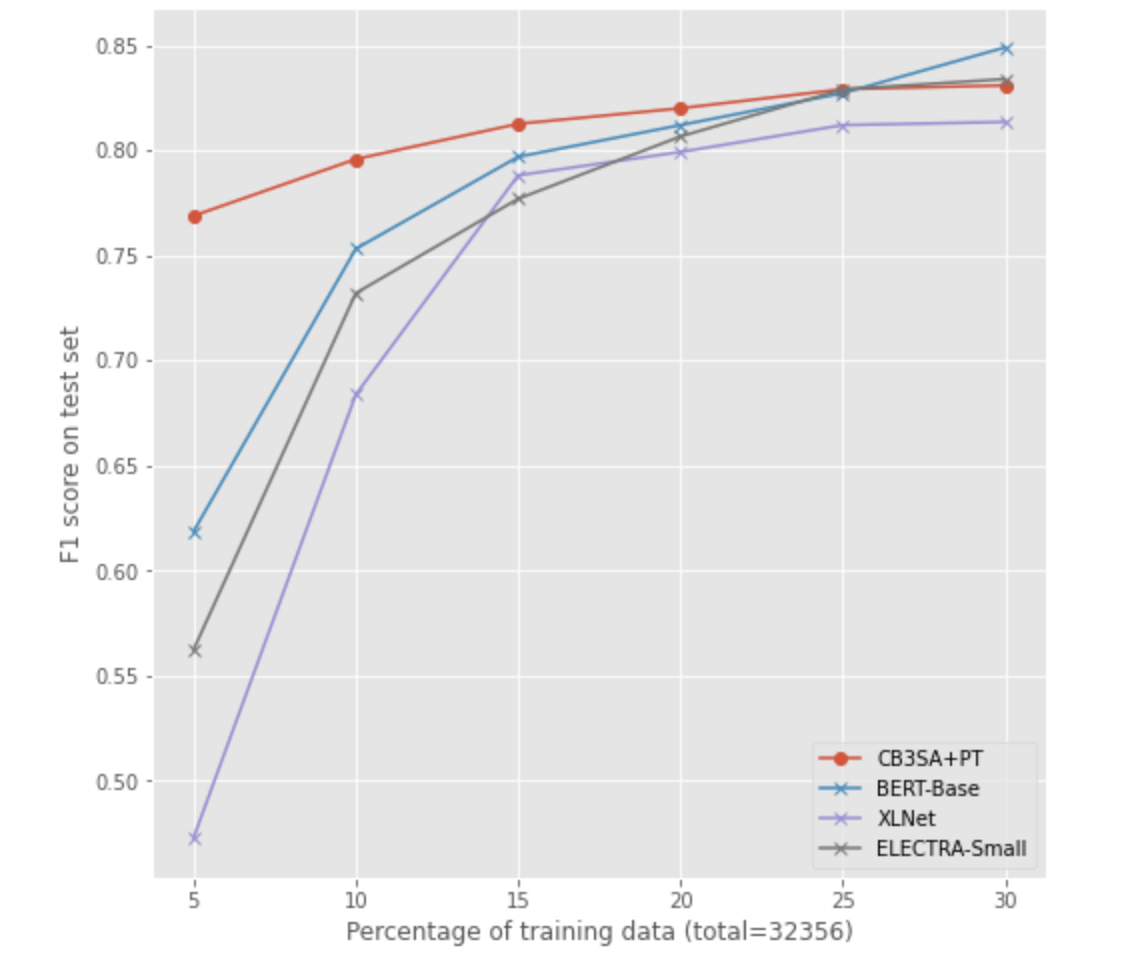}
  \caption{F1 score on test set when allowed to fine tune only on a small fraction of training data. Parameter heavy state-of-the-art models rapidly deteriorate as amount of training data is reduced while our pre trained model is resilient in the low data regime.}
  \Description{Parameter heavy state-of-the-art models rapidly deteriorating as amount of training is being reduced while our pre trained model is resilient in the low data regime.}
  \label{low:resource}
\end{figure}
We experiment in a low resource setting by allowing all the models to be fine tuned only on a small fraction of the training data and then evaluating their performance on the same test dataset as was used in Section 6.3. Figure \ref{low:resource} compares the performance of our pre trained model with that of the SOTA models which shows our model's performance deteriorates much lesser compared to others. \textbf{When allowed to be fine tuned only on 5\% of the training data (around 1600 parallel examples) our model outperforms the best performing SOTA model, BERT-Base, by 24.3\%}. In fact, our model continues outperforming all SOTA models for a large part of the low data regime, till 20\%, after which others catch up. 

\begin{table}
  \caption{Sample title compression by CB3SA+PT}
  \label{tab:result}
  \begin{tabular}{p{0.31\textwidth}p{0.10\textwidth}}
    \toprule
    Long Title&Short Title\\
    \midrule
    Glad OdorShield Tall Kitchen Drawstring Trash bags - Febreze Fresh Clean - 13 Gallon - 40 count (Packaging May Vary) & trash bags\\
    \midrule
    Great Value Strawberry Nonfat Greek Yogurt, 6 oz, 4 ct & nonfat greek yogurt\\
    \midrule
    Del Monte Fresh Cut Cut Green Beans \& Potatoes With Ham Style Flavor, 29 Oz & green beans and potatoes\\
    \midrule
    Suave Professionals Moisturizing Shampoo and Conditioner Almond + Shea Butter 28 oz, 2 count&shampoo and conditioner\\
  \bottomrule
\end{tabular}
\end{table}

\section{Conclusion}
We propose a self-attentive recurrent neural network architecture for product title compression. We successfully pre train our network as a discriminator on a replaced token detection task on unlabeled dataset of long product titles. We also fine tune several large state of the art pre trained NLP models for the title summarization task. Our experiments with human generated short titles on the Walmart grocery catalog show that our pre trained model with 2 Million parameters achieves performance comparable to to state of the art models like BERT and XLNet that use in excess of 100 Million parameters. Further, in a low resource setting when very small amount of training data is available, our model outperforms all SOTA models by a large margin. 

\bibliographystyle{ACM-Reference-Format}
\bibliography{sigirecomm}

\end{document}